\documentclass[nohyperref]{article}

\usepackage{microtype}
\usepackage{graphicx}
\usepackage{subfigure}
\usepackage{booktabs} 

\usepackage{hyperref}

\usepackage[accepted]{icml2023}

\usepackage{amsmath}
\usepackage{amssymb}
\usepackage{mathtools}
\usepackage{amsthm}
\usepackage{listings}

\usepackage[capitalize,noabbrev]{cleveref}

\definecolor{codegreen}{rgb}{0,0.6,0}
\definecolor{codegray}{rgb}{0.5,0.5,0.5}
\definecolor{codepurple}{rgb}{0.58,0,0.82}
\definecolor{backcolour}{rgb}{0.97,0.97,0.97}

\lstdefinestyle{mystyle}{
    backgroundcolor=\color{backcolour},
    commentstyle=\color{codegreen},
    keywordstyle=\color{magenta},
    numberstyle=\tiny\color{codegray},
    stringstyle=\color{codepurple},
    basicstyle=\ttfamily\footnotesize,
    breakatwhitespace=false,
    breaklines=true,
    captionpos=b,
    keepspaces=true,
    numbers=left,
    numbersep=5pt,
    showspaces=false,
    showstringspaces=false,
    showtabs=false,
    tabsize=2
}

\theoremstyle{plain}

\theoremstyle{definition}

\theoremstyle{remark}

\usepackage[textsize=tiny]{todonotes}

\icmltitlerunning{Time Interpret}

\begin{document}

\twocolumn[
\icmltitle{Time Interpret: a Unified Model Interpretability Library for Time Series}



\icmlsetsymbol{equal}{*}

\begin{icmlauthorlist}
\icmlauthor{Joseph Enguehard}{babylon,skippr}
\end{icmlauthorlist}

\icmlaffiliation{babylon}{Babylon Health, 1 Knightsbridge Grn, London SW1X 7QA United Kingdom}
\icmlaffiliation{skippr}{Skippr, 99 Milton Keynes Business Centre, Milton Keynes MK14 6GD United Kingdom}
\icmlcorrespondingauthor{Joseph Enguehard}{joseph@skippr.com}

\icmlkeywords{Machine Learning}

\vskip 0.3in
]



\printAffiliationsAndNotice{}  

\begin{abstract}
    We introduce \texttt{\detokenize{time_interpret}}, a library designed as an extension of
Captum~\citep{kokhlikyan2020captum}, with a specific focus on temporal data.
As such, this library implements several feature attribution methods that can be used to explain predictions made by
any Pytorch model.
\texttt{\detokenize{time_interpret}} also provides several synthetic and real world time series datasets, various PyTorch
models, as well as a set of methods to evaluate feature attributions.
Moreover, while being primarily developed to explain predictions based on temporal data, some of its components have a
different application, including for instance methods explaining predictions made by language models.
In this paper, we give a general introduction of this library.
We also present several previously unpublished feature attribution methods, which have been developed along with
\texttt{\detokenize{time_interpret}}\footnote{
    This library can be found at \url{https://github.com/josephenguehard/time_interpret}.
}.
\end{abstract}

\section{Introduction}
\label{sec:introduction}

As neural networks are becoming more relied on for many decision processes, there has been an increased focus into
understanding how these algorithms come to a specific output.
Should such models be used to take high-stakes decisions, as it is often the case in finance or in medicine, it is
indeed crucial for these tools to provide a justification along with their results, which could be used to discuss
or even oppose a decision.
The risk is that deep learning models, often considered as $``$black boxes$"$, could produce biased and unfair behaviour.
Such behaviour has already been noticed in opaque algorithms such as COMPAS, used to evaluate the risk of recidivism,
which was accused of being biased toward people of color~\citep{rudin2019stop}.

As a result, many methods aiming to explain a deep learning prediction have been developed, which now constitutes a
field usually called Explainable AI (XAI).
Among these methods, some of the most common are LIME~\citep{ribeiro2016should}, SHAP~\citep{lundberg2017unified},
or Integrated Gradients~\citep{sundararajan2017axiomatic}.\footnote{
    An insightful presentation of these methods can be found at \url{https://christophm.github.io/interpretable-ml-book}.
}
Alongside this research, there has also been a drive to create libraries unifying these different methods, enabling
their use on popular deep learning libraries, as well as integrating evaluation tools to compare these XAI methods.
To this end, several pieces of software have been proposed, including SHAP~\citep{lundberg2017unified},
InterpretML~\citep{nori2019interpretml}, OmniXAI~\citep{wenzhuo2022-omnixai} or Captum~\citep{kokhlikyan2020captum}
, among many.

However, several researchers~\citep{tonekaboni2020went, crabbe2021explaining} have noticed a lack of attention toward
a specific type of data: time series.
Temporal data is nevertheless crucial in many applications: indeed, for instance, financial and medical data commonly
consist in multivariate time series.
Models which produce predictions based on this type of data therefore need a careful consideration, as these
applications often carry high-stakes decisions.
Consequently, several feature attribution methods have recently been introduced to tackle this specific
case~\citep{choi2016retain, tonekaboni2020went, crabbe2021explaining}.
Yet, to the best of our knowledge, there seems to be a lack of a unified library to regroup and evaluate these specific
methods.\footnote{
    With the exception of TSInterpret~\citep{hollig2022tsinterpret}, another library which has this specific focus.
    We concurrently developed \texttt{\detokenize{time_interpret}} unaware of this work, which contains several methods
    not covered by our library.
    Therefore, we also recommend this library to the reader.
}

As a result, we created \texttt{\detokenize{time_interpret}} (short: \texttt{tint}), a Python library designed as an
extension of Captum~\citep{kokhlikyan2020captum}.
Although this library can be used with any PyTorch~\citep{NEURIPS2019_9015} model, it has a specific focus on time series,
providing several feature attribution methods developed for this specific type of data.
\texttt{\detokenize{time_interpret}} also provides evaluation tools, whether the true attributions are known or not, as
well as several time series datasets.
It also leverages PyTorch Lightning~\citep{Falcon_PyTorch_Lightning_2019} to simplify the use of the original PyTorch
library.
As such, it provides several common PyTorch models used to handle temporal data, as well as a specific PyTorch Lightning
wrapper.

Moreover, despite this focus on time series, several components of \texttt{\detokenize{time_interpret}} have a slightly
different application.
It provides for instance various methods aiming to explain language models such a BERT\@.
Its evaluation tools can also be used with any feature attribution methods, and not just the ones implemented in
this library.

This paper aim to give a general introduction to \texttt{\detokenize{time_interpret}}.
Furthermore, several previously unpublished methods have been developed along with this library, which we also present
here.
We hope this study will give more clarity to the corresponding codebase, and will prove useful for further research
in this field.
We encourage the reader to also refer to the library documentation for more information, especially in case of
new significant releases.
\section{Presentation of the library}
\label{sec:presentation}

We provide in this section an introduction to the \texttt{\detokenize{time_interpret}} library.
Please also refer to the documentation.

\texttt{\detokenize{time_interpret}} is primarily composed of 4 different parts: attribution methods, datasets,
evaluation tools (metrics) and deep learning models.
We present below a short description of the components in each of these parts.

\paragraph{Attribution methods.}

Attribution methods constitute the core of \texttt{\detokenize{time_interpret}}.
In this part of the library, we regrouped many methods which have been recently published.
Similarly to Captum~\citep{kokhlikyan2020captum}, each method can be called like this:

\begin{lstlisting}[language=Python, caption=Attribution loading example, label={lst:attr}]
from tint.attr import TemporalIntegratedGradients

explainer = TemporalIntegratedGradients(
    model
)
attr = explainer.attribute(inputs)
\end{lstlisting}

where $``$model$"$ is a PyTorch model, and $``$inputs$"$ is an inputs' tensor.

We provide in this library several methods:

\begin{itemize}
    \item \textbf{AugmentedOcclusion}.
        This method improves upon the original Occlusion method from captum~\url{https://captum.ai/api/occlusion.html}
        by allowing to sample the baseline from a bootstrapped distribution.
        By selecting a distribution close to the inputs, the resulting occulted data should be close to actual data,
        limiting the amount of out of distribution samples.
        This method was originally proposed by~\citep{tonekaboni2020went}, Section 4.
    \item \textbf{BayesLime, BayesKernelShap}.
        These two methods, originally proposed by~\citep{slack2021reliable}, extend respectively
        LIME~\citep{ribeiro2016should} and KernelSHAP~\citep{lundberg2017unified}, by replacing the underlying
        linear regression model with a bayesian linear regression, allowing the method to model uncertainty in
        explainability by outputting credible intervals in addition to the feature attributions.
    \item \textbf{DiscretetizedIntegratedGradients (DIG)}.
        DIG~\citep{sanyal2021discretized} was designed to interpret predictions made by language models.
        It builds upon the original Integrated Gradients method by generating discretized paths, hopping from one
        word to another, instead of using straight lines.
        This way, it aims to create a path which takes into account the discreteness of the embedding space.
    \item \textbf{DynaMask}.
        This method, introduced by~\citep{crabbe2021explaining}, is an adaptation of a perturbation-based method
        developed in~\citep{fong2017interpretable, fong2019understanding}, to handle time-series data.
        As such, it consists in perturbing a temporal data by replacing some of it with an average in time.
        The mask used to choose which data should be preserved and which should be replaced is learnt in order to either
        preserve the original prediction with a minimum amount of unmasked data, or change the original prediction with
        a small amount of masked data.
        Either way, the learnt mask can then be used to discriminate between important features and others.
    \item \textbf{ExtremalMask}.
        This method~\citep{enguehard2023learning} consists in a generalisation of DynaMask, which learns not only the
        mask, but also the associated perturbation, instead of replacing perturbed data with a predetermined average in
        time.
        Learning perturbations allows this method to take into account eventual long term dependencies, such as
        temporal regularities.
    \item \textbf{Fit}.
        Originally proposed by~\citep{tonekaboni2020went}, this method aims to understand which feature is important by
        quantifying the shift in the predictive distribution over time.
        An important feature is then one which contributes significantly to the distributional shift.
    \item \textbf{LofLime, LofKernelShap}.
        Novel method.
        Please see Section~\ref{sec:methods} for more details.
    \item \textbf{NonLinearitiesTunnel}.
        Novel method.
        Please see Section~\ref{sec:methods} for more details.
    \item \textbf{Retain}.
        This method~\citep{choi2016retain} uses two RNNs whose outputs are then used as keys and queries in an attention
        mechanism, also using the original embeddings of the input as values.
        This attention mechanism can then be used to explain which feature was important to make a specific prediction.
    \item \textbf{SequentialIntegratedGradients (SIG)}.
        SIG~\citep{enguehard2023sequential} is an adaptation of the Integrated Gradients method to sequential data.
        It modifies the baseline by only masking one element of a sequence at a time, and computing the corresponding
        feature attribution.
        By doing so, it allows the baseline to be closer to the original sequence.
    \item \textbf{TemporalOcclusion}.
        Originally proposed by~\citep{tonekaboni2020went}, this method modifies the Occlusion method from
        Captum~\url{https://captum.ai/api/occlusion.html} by only masking the last input data in time, preserving the
        previous inputs.
    \item \textbf{TemporalAugmentedOcclusion}.
        This method combines TemporalOcclusion and AugmentedOcclusion.
        As such, it only masks the last input in time, replacing it with samples from a bootstrapped distribution.
    \item \textbf{TemporalIntegratedGradients}.
        Novel method.
        Please see Section~\ref{sec:methods} for more details.
    \item \textbf{TimeForwardTunnel}.
        Novel method.
        Please see Section~\ref{sec:methods} for more details.

\end{itemize}

\paragraph{Datasets}

As part of \texttt{\detokenize{time_interpret}}, we include a collection of datasets which can be readily used:

\begin{lstlisting}[language=Python, caption=Dataset loading example, label={lst:datasets}]
from tint.datasets import Arma

arma = Arma()
arma.download()  # This method generates the dataset

inputs = arma.preprocess()["x"]
true_saliency = arma.true_saliency(dim=1)
\end{lstlisting}

All these datasets are implemented using the DataModule from PyTorch Lightning.
We provide the following datasets:

\begin{itemize}
    \item \textbf{Arma}.
        This dataset was introduced by~\citep{crabbe2021explaining}.
        It relies on an ARMA process to generate time features $x_{ti}$, which are inputs combined by white-box
        regressor f.
        This regressor uses only part of the data, so we can know in advance which feature is salient, in order to
        evaluate a features attributions method.
        As a result, this dataset can be used to provide an initial comparison of various methods on time-series data.
    \item \textbf{BioBank}.
        BioBank~\citep{sudlow2015uk} is a large dataset based on the UK population.
        As such, it can be used to predict a specific condition based on a number of patients' data.
        In the default setting, this dataset can be used to train a model to predict the risk of developing type-II
        diabetes, but other conditions could be predicted.
        We also provide a cutoff before the onset of the condition, only using features before that time, as well as
        a script to train FastText~\citep{bojanowski2017enriching} embeddings on the medical codes.
        The dataset can also be discretized, grouping medical codes into time intervals such as years.
        To access the data, a formal application must be made, please see
        \url{https://www.ukbiobank.ac.uk/enable-your-research}.
    \item \textbf{Hawkes}.
        Hawkes processes are a specific type of temporal point processes (TPP).
        The probability of an event of type k happening at time t is conditioned by a specific intensity function:
        \[ \lambda^*_k(t) = \mu_k + \sum_{n=1}^K \alpha_{kn} \sum_{t_i^n < t} \exp \left[ -\beta_{kn} (t - t_i^n) \right] \]
        with $\bm{\alpha}$, $\bm{\beta}$ and $\bm{\mu}$ being parameters of the process.
        As a result, an event has greater chance of happening if an event already happened previously, depending on the
        values of $\bm{\alpha}$ and $\bm{\beta}$.
        If no event previously happened, the base probability of an event happening is determined by $\bm{\mu}$.
        In this case, this intensity can therefore be used as the true saliency of the features.
        An important consequence for this dataset is that the true saliency is \textbf{dependent in time}.
        Indeed, an event can be very important for another one close in time, but also irrelevant for other events.
        As a result, the true saliency has two temporal dimensions: for each time, it gives the importance of each
        temporal event.
    \item \textbf{Hidden Markov model (HMM)}.
        This dataset was introduced by~\citep{crabbe2021explaining}.
        It consists in a 2-state hidden Markov model, with a features vector of size 3.
        For each time, only one of the three features conditions the label: either the second or the third, depending
        on the state.
        The first feature is always not salient.
        For this dataset, the true salient features are therefore also known.
    \item \textbf{Mimic-III}.
        Mimic-III~\citep{johnson2016mimic} consists of ICU data points.
        This dataset provides two tasks: predicting in-hospital mortality (static label) and predicting the
        average blood pressure (temporal label).
        The processing of this dataset was done following~\citep{tonekaboni2020went, crabbe2021explaining}.
\end{itemize}

\newpage

\paragraph{Metrics}

Metrics are an important component of \texttt{\detokenize{time_interpret}}, as they provide elements of comparison
between feature attribution methods in two scenarios: when the true salient features are known, and when they are
unknown.

Both metrics can be imported this way:

\begin{lstlisting}[language=Python, caption=Attribution evaluation example, label={lst:metrics}]
from tint.metrics import accuracy
from tint.metrics.white_box import aup

print(f"{aup(attr, true_saliency):.4}")
print(f"{accuracy(model, input, attr):.4}")
\end{lstlisting}

As we can see on Snippet~\ref{lst:metrics}, these metrics have a different behavior in each scenario.
In the case where the true salient features are known, the metrics are straightforward: they directly compare some
attributions with the truth.
\texttt{\detokenize{time_interpret}} provides the following metrics: area under precision (AUP), area under recall (AUR),
area under precision recall curve (AUPRC), Roc-Auc, mean average error (MAE), mean square error (MSE), root mean square
error (RMSE).
Following~\citep{crabbe2021explaining}, we also provide the Entropy and Information metrics.

In the case where the true salient features are not known, we draw from the work
of~\citep{shrikumar2017learning,deyoung2019eraser,crabbe2021explaining} and propose a list of metrics following a
similar pattern.
Each metric is computed by masking the top or bottom x\% most important features.
We then compute predictions by the model to be explained on this masked data.
The shift in the resulting predictions provides information on how important the masked data actually was.

Moreover, in order to avoid creating out of distribution data by masking some of the features, we allow the user to
pass a custom baseline to the metric.
It is also possible to draw baselines from a distribution, to add gaussian noise to the masked inputs, and to compute
predictions multiple times to produce consistent results.
In the last case, a batch size can be passed to compute these predictions in parallel and reduce time complexity.

Furthermore, it is also possible to pass a weight function to the metric: the shift in prediction given some masked
data influences the metric proportionally to some weight.
We provide two weights functions: $``$lime-weight$"$ weights the results according to a cosine or euclidean distance
between the masked data and the original one.
$``$lof-weight$"$ weights the results according to a Local Outlier Factor score, to reduce the influence of potential
outliers.

\newpage

We provide the following metrics:

\begin{itemize}
    \item \textbf{accuracy}.
        This metric measures by how much the accuracy of a model drops when
        removing or only keeping the topk most important features.
        Lower is better.
        A custom threshold can also be passed to discriminate between positive and negative predictions.
    \item \textbf{comprehensiveness}.
        This measures by how much the predicted class probability changes when removing the topk most important
        features.
        Higher is better.
    \item \textbf{cross-entropy}.
        This metric measures the cross-entropy between the outputs of the model using the original inputs and perturbed
        inputs by removing or only keeping the topk most important features.
        Higher is better.
    \item \textbf{Log-odds}.
        This log-odds measures the average difference of the negative logarithmic probabilities on the predicted class
        when removing the topk most important features.
        Lower is better.
    \item \textbf{mae}.
        This metric measures the mean absolute error between the outputs of the model using the original inputs and
        perturbed inputs by removing or only keeping the topk most important features.
        Lower is better.
    \item \textbf{mse}.
        This metric measures the mean square error between the outputs of the model using the original inputs and
        perturbed inputs by removing or only keeping the topk most important features.
        Lower is better.
    \item \textbf{sufficiency}.
        This sufficiency measures by how much the predicted class probability changes when keeping only the topk most
        important features.
        Lower is better.
\end{itemize}

Finally, we provide an additional metric, $``$lipschitz-max$"$, measuring the stability of a feature attribution method
on small variations of the input data.
However, sensitivity-max should be preferred, as a more robust method.
Please see \url{https://captum.ai/api/metrics.html#sensitivity} for more details.
We still provide it as it has been used in some research and could be useful to reproduce experiments.

\paragraph{Models}

We provide several PyTorch models: MLP, CNN, RNN and TransformerEncoder, as well as some language models
from HuggingFace: Bert, DistilBert and RoBerta.
These models can be easily wrapped into a PyTorch-lightning model as such:

\begin{lstlisting}[language=Python, caption=Model definition example, label={lst:model}]
from tint.models import MLP, Net

mlp = MLP(units=[5, 10, 1])
net = Net(
    mlp, loss="cross_entropy", optim="adam"
)
\end{lstlisting}

\section{Novel methods}
\label{sec:methods}

\begin{figure*}[t]
\begin{center}
\centerline{\includegraphics[width=0.6\textwidth]{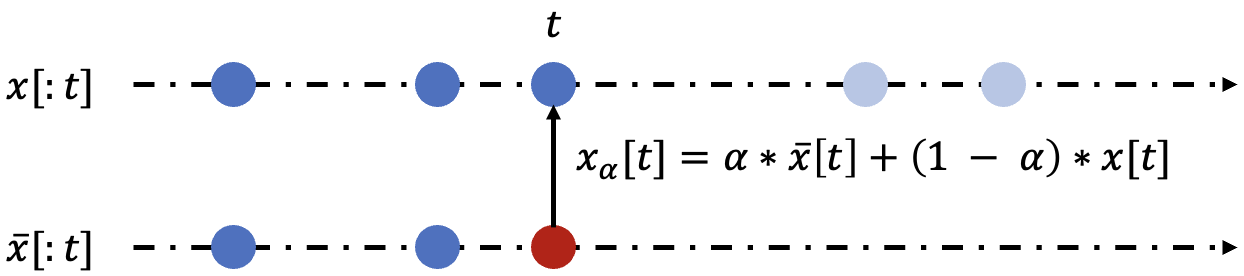}}
\caption{
    \textbf{Illustration of the TIG method}.
    Given an input $\textbf{x}$ and a corresponding baseline $\overline{\textbf{x}}$, we compute the attribution as such.
    For each time t, we crop future times: $\textbf{x}[:t]$, and we only compute interpolations on the last data point in
    time: $\textbf{x}_{\alpha} = (x_1, x_2, \dots, \alpha \times \overline{\textbf{x}}[t] + (1 - \alpha) \times \textbf{x}[t])$.
    Using these interpolated points, we can then compute the integrated gradients.
}
\label{fig:tig}
\end{center}
\end{figure*}

We present in this section several previously unpublished methods, which have been developed
with \texttt{\detokenize{time_interpret}}.
There are two types of such methods: ones which are specific to temporal data, and others which are more general.

\subsection{Temporal attribution methods}
\label{subsec:temporal-attribution-methods}

\paragraph{Temporal Integrated Gradients.}

Temporal Integrated Gradients (TIG), illustrated on Figure~\ref{fig:tig}, adapts the original Integrated Gradients
method~\citep{sundararajan2017axiomatic} to specifically handle time-series data.
In this method, instead of directly computing attribution by summing gradients over interpolated points between the
original data $\textbf{x}$ and an uninformative baseline $\overline{\textbf{x}}$, we first crop the temporal sequence up
to a specific time t.
Then, we keep all the data points in time fixed, only computing interpolations on the last data point at time t.
The baseline of the sequence $\textbf{x} \in \mathbb{R}^{TN}$, where $T$ is the temporal dimension and $N$ the
feature dimension, can therefore be defined as:

\[ \overline{\textbf{x}}[:t] = (\textbf{x}_1, \dots, \textbf{x}_{t-1}, \overline{\textbf{x}}_t) \]

Using this baseline, we can now define TIG on time t and feature i:

\begin{equation}
    \textrm{TIG}_{ti} \coloneqq  (x_{ti} - \overline{x}_{ti}) \times
    \int_0^1 \frac{\partial \textrm{F}(\overline{\textbf{x}}[:t] + \alpha \times (\textbf{x}[:t] -
    \overline{\textbf{x}}[:t]))}{\partial x_{ti}} \, d\alpha
    \label{eq:tig}
\end{equation}

The overall attribution of a temporal point $\textbf{x}_t$ is then computed by summing over the feature dimension i and
normalising the result:

\[ \textrm{TIG}_t(\textbf{x}) \coloneqq \frac{\sum_i \textrm{TIG}_{ti}}{||\textrm{TIG}||} \]

As a result, in this setting, we keep the baseline as close as possible to the original input data, only modifying the
last input in time.
Moreover, we remove any future data points, so that they do not influence the attribution of the data point $\textbf{x}_t$.
This method can also be seen as an adaptation of SIG~\citep{enguehard2023sequential}, applied not on sequences, but on
temporal data.

\paragraph{Time Forward Tunnel.}

This method extends the one presented in~\citep{tonekaboni2020went}, allowing any Captum method to be used on temporal
data.
As such, this tool computes attribution iteratively by cropping the input data $\textbf{x} \in \mathbb{R}^{TN}$ up to
a time t, and performing the passed feature attribution method:

\[ \textrm{TimeForwardTunnel(AttrMethod)}(\textbf{x}_t) = \textrm{AttrMethod}(\textbf{x}[:t])_t \]

As a result, future data is not used when computing the attribution of a temporal data $\textbf{x}_t$.

Moreover, instead of passing a target, Time Forward Tunnel allows to use a model's predictions at time t as the target,
supporting a range of tasks: binary, multilabel, multiclass and regression ones.

Furthermore, note that TimeForwardTunnel used with IntegratedGradients is different from TIG, as the former method uses
$(\overline{\textbf{x}}_1, \dots, \overline{\textbf{x}}_t)$ as a baseline, while TIG uses
$(\textbf{x}_1, \dots, \textbf{x}_{t-1}, \overline{\textbf{x}}_t)$,
only computing interpolations on the last temporal data point.

\paragraph{Temporal attributions, targets and additional arguments.}

A number of implemented method support $``$temporal attribution$"$, meaning that, in this setting, they will return
feature attributions at each temporal point.
Therefore, if the input data is of shape $\textbf{x} \in \mathbb{R}^{TN}$, the corresponding attribution will be of
shape $\textrm{Attr} \in \mathbb{R}^{TTN}$, with an extra dimension $T$ representing the temporality of the attributions.
For instance, $\textbf{x}_t$ could be important at time $t_1$, $\textrm{Attr}_{t_1}$ having high values, but could also
be less important at a later time $t_2$, $\textrm{Attr}_{t_2}$ having lower values.

The supported methods currently are:

\begin{itemize}
    \item DynaMask
    \item ExtremalMask
    \item Fit
    \item Retain
    \item Temporal IG
    \item Time Forward Tunnel
\end{itemize}

\subsection{Other attribution methods}
\label{subsec:other-attribution-methods}

\paragraph{Local Outlier Factor (LOF) LIME and Kernel SHAP\@.}

LOF LIME and LOF Kernel SHAP are variations of the vanilla methods LIME and SHAP, using the Local Outlier Factor
score~\citep{breunig2000lof} as an indication of similarity between the masked data and the original one:

\[ \textrm{SimilarityScore}(\textbf{x}; X) = \frac{1}{\max(1, \textrm{LOF}(\textbf{x}, X))} \]

where \textbf{x} is a data point to be explained, and $X$ is a set of data points.
A low value of LOF indicates an inlier, while a large value an outlier.
As a result, SimilarityScore is lower when \textbf{x} is an outlier.

This similarity measure is then used as a weight of the interpretable sample for training an interpretable model.
Therefore, with these methods, it matters less how different a masked data point is, compared with the original one.
What matters here is how far the masked data point is from a data distribution.
These methods should therefore be less sensitive to outliers.
However, to properly work, they require a number of data points to be used as the original data distribution.

\paragraph{Non-linearities tunnel.}

This method expands the work of~\citep{dombrowski2019explanations}, by replacing some of a model's modules before
computing feature attributions.
By default, this tool replaces ReLU activations with Softplus ones, in order to reduce the potential to manipulate
explanations due to a predictive space with a high curvature, as explored by~\citep{dombrowski2019explanations}.
However, any module could similarly be replaced here.

\section{Conclusion}
\label{sec:conclusion}

We have presented in this study \texttt{\detokenize{time_interpret}} an extension of Captum with a specific focus on
time series, along with several novel attribution methods.
This repository also comes with a number of experiments, using methods implemented in this library.
We did not report the results of these experiments in this paper, but the reader can refer to them in the relevant
sections of the codebase.
We hope that this work will foster further research in the field of explainable AI applied to time series.


\bibliography{bibliography}
\bibliographystyle{icml2023}


\end{document}